\documentclass{article}

    \PassOptionsToPackage{numbers, compress}{natbib}

\usepackage[preprint]{neurips_2024}

\usepackage[utf8]{inputenc} 
\usepackage[T1]{fontenc}    
\usepackage{hyperref}       
\usepackage{url}            
\usepackage{booktabs}       
\usepackage{amsfonts}       
\usepackage{nicefrac}       
\usepackage{microtype}      
\usepackage{xcolor}         
\usepackage{amsmath,amsthm,amssymb}
\usepackage{cleveref}

\usepackage{graphicx}

\title{Evidence from fMRI Supports a Two-Phase Abstraction Process in Language Models}

\author{%
  Emily Cheng$^*$ \\
  Universitat Pompeu Fabra\\
  Barcelona, Spain \\
  \texttt{emilyshana.cheng@upf.edu} \\
  \And
  Richard J. Antonello$^*$ \\
  Columbia University \\
  New York City, New York \\
  \texttt{rja2163@columbia.edu} \\
}

\begin{document}

\def\thefootnote{*}\footnotetext{These authors contributed equally to this work.}\def\thefootnote{\arabic{footnote}}

\maketitle

\begin{abstract}
Research has repeatedly demonstrated that intermediate hidden states extracted from large language models are able to 
predict measured brain response to natural language stimuli. Yet, very little is known about the 
representation properties that enable this high prediction performance. Why is it the intermediate layers, and not the output layers, that are most capable for this unique and highly general transfer task? In this work, we show
that evidence from language encoding models in fMRI supports the existence of a 
two-phase abstraction process within LLMs. We use manifold learning methods to show that this abstraction process naturally arises over the course of training a language model and that the first "composition" phase of this abstraction process is compressed into fewer layers as training continues. Finally, we demonstrate a strong correspondence between layerwise encoding performance and the intrinsic dimensionality of representations from LLMs. We give initial evidence that this correspondence primarily derives from the inherent compositionality of LLMs and not their next-word prediction properties. \end{abstract}

\section{Introduction}
How do brains and machines take low-level information, such as a collection of sounds or words, and compose it into the rich tapestry of ideas and concepts that can be expressed in natural language? This question of composition, or abstraction, is at the heart of most studies of human language comprehension. Recent work has shown that representations from large language models (LLMs) are able to successfully model human brain activity at varying spatial and temporal resolutions with only a linear transformation~\cite{goldstein2022shared, vaidya2022self, jain2023computational, NEURIPS2023_4533e4a3, tuckute2023driving, NEURIPS2023_3a0e2de2, mischler2024contextual}. This has led to questions about the reason for this brain-model similarity. Do LLMs and brains possess similar representations because they have similar learning properties or objectives?~\cite{caucheteux2023evidence, schrimpf2021neural, goldstein2022shared} Or is the similarity merely a consequence of shared abstraction, the ability to represent features not derivable from the lexical properties of language alone?~\cite{antonello2022predictive} 

In this work, we present new evidence that it is the abstractive, compositional properties of LLMs that drive predictivity between LLMs and brains. We do this by examining an underexplored and unexplained phenomenon of the similarity - the tendency for intermediate hidden layers of LLMs to be optimal for this linear transfer task. We show that an LLM layer's performance at predicting brain activity is strongly related to intrinsic dimensionality of that layer relative to other layers in the same network. Furthermore, we demonstrate that this relationship is itself an indicator that pretrained LLMs naturally split into an early \emph{abstraction}, or composition, phase, and a later \emph{prediction}, or extraction, phase, a result independently suggested in the LM interpretability literature \citep{valeriani2023,cheng2024emergencehighdimensionalabstractionphase}. We suggest that it is the first abstraction phase, rather than the latter prediction phase, that primarily drives the observed correspondence between brains and LLMs. 

\section{Methods}
We test the hypothesis that feature abstraction, not next-token prediction \emph{per se}, drives brain-model similarity. To do so requires three observables. First, we measure the dependent variable, \textbf{(1)} brain-model representational similarity, by scoring the prediction performance of a learned linear mapping from LLM representations to brain activity. Then, we compute the \textbf{(2)} dimensionality of representations to measure abstract feature complexity over the LM's layers. Finally, to test the alternate hypothesis that next-token prediction drives brain-LM similarity, as has been suggested by others~\cite{schrimpf2018brain, caucheteux2023evidence, goldstein2022shared}, we compute the \textbf{(3)} \emph{surprisal}, or next-token prediction error, from each layer. In contrast to prior surprisal measurement approaches \cite{antonello2022predictive}, we compute this layerwise surprisal using the TunedLens approach devised by \citet{Belrose2023ElicitingLP} to reduce measurement noise.

\subsection{Brain-model similarity}

\paragraph{fMRI data}


We used publicly available functional magnetic resonance imaging (fMRI) data collected from 3 human subjects as they listened to 20 hours of English language podcast stories over Sensimetrics S14 headphones. Stories came from podcasts such as \textit{The Moth Radio Hour}, \textit{Modern Love}, and \textit{The Anthropocene Reviewed}. Each 10-15 minute story was played during a separate scan. Subjects were not asked to make any responses, but simply to listen attentively to the stories. For encoding model training, each subject listened to roughly 95 different stories, giving 20 hours of data across 20 scanning sessions, or a total of   \textasciitilde33,000 datapoints for each voxel across the whole brain. Additional details of the MRI methods are summarized in Appendix \ref{app:fmri}.

\paragraph{Neural encoding model training}

To train encoding models, we use the method described in \cite{NEURIPS2023_4533e4a3}. For each word in the stimulus set, activations were extracted by feeding that word and its immediate preceding context into the LLM. A sliding window was used to ensure each word received a minimum of 256 tokens of context. Activations were then downsampled using a Lanczos filter and FIR delays of 1,2,3 and 4 TRs were added to account for the hemodynamic lag in the BOLD signal. A linear projection from the downsampled, time-delayed features was trained using ridge regression. Encoding models were built using the OPT language model \cite{Zhang_Roller_Goyal_Artetxe_Chen_Chen_Dewan_Diab_Li_Lin_et} (three sizes - 125M, 1.3B, 13B) and the 6.9B parameter deduped Pythia language model \cite{biderman2023pythia}. To study model training, 9 different Pythia model checkpoints were used (at 1K, 2K, 3K, 4K, 8K, 16K, 32K, 64K, and 143K training steps). 

\begin{figure}
    \centering
    \includegraphics[width=\linewidth]{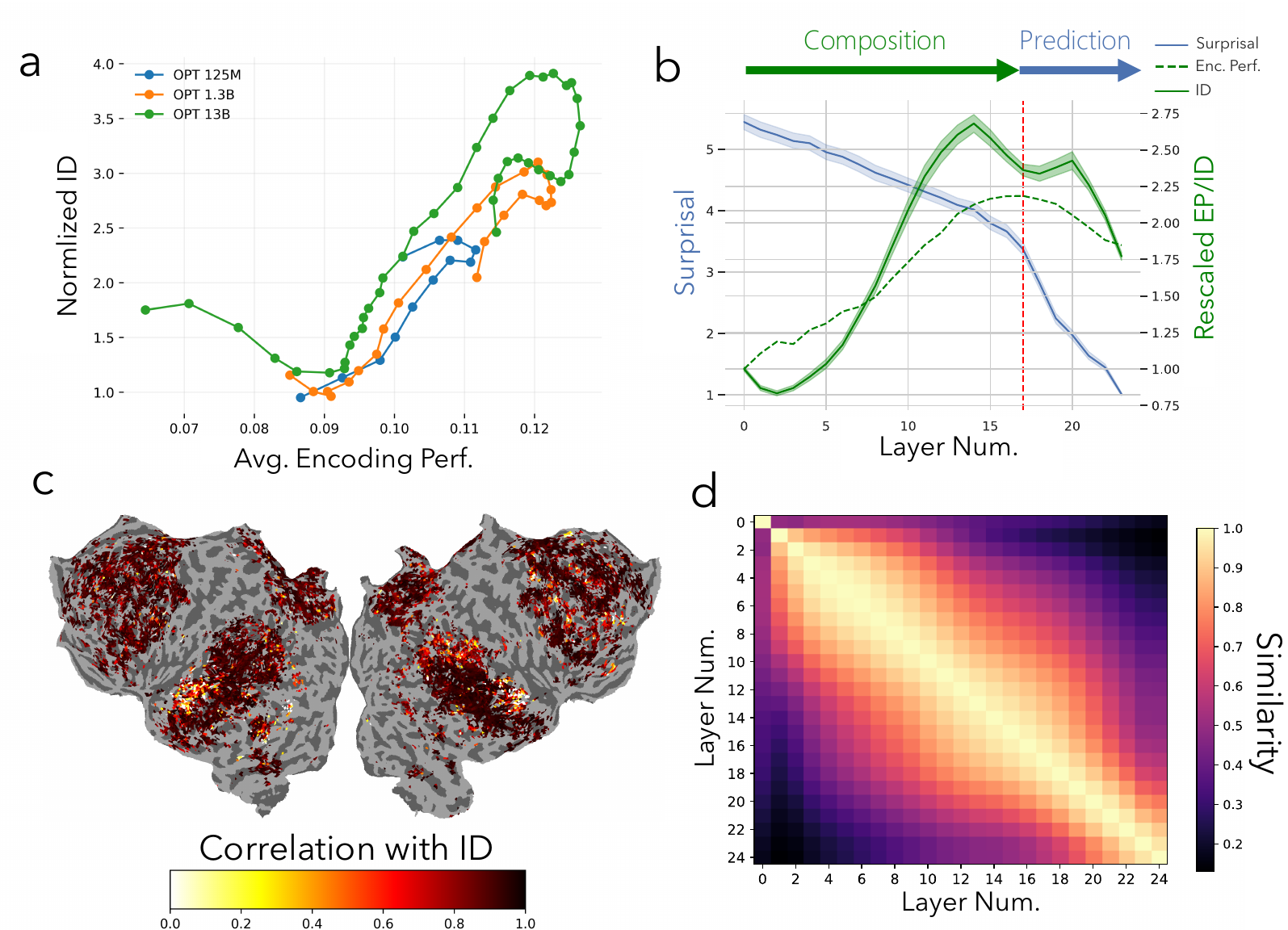}
    \caption{\textit{Analyzing Layerwise Representational Trends}: \textbf{(a)} $I_d$ is well correlated with encoding performance across model sizes. $I_d$ is normalized here by the log of embedding size to account for power law scaling. \textbf{(b)} The abstract-predict phase transition at layer 17 is shown for OPT-1.3b. At the peak of encoding performance (red dashed line), the next-token prediction loss (blue curve) sharply decreases, corresponding with a decrease in encoding performance. \textbf{(c)} A flatmap of the brain, for one subject, is shown colored voxelwise by the correlation over layers between $I_d$ and encoding performance. With the exception of auditory cortex (bright), which captures low-level spectral information, encoding performance in brain regions thought to perform higher-level linguistic processing (dark) is well-captured by representational $I_d$.\textbf{(d)} The layer-wise representational similarity computed with linear CKA is shown for OPT-1.3B.}
    \label{fig:phase_transition_lm}
\end{figure}

\subsection{Dimensionality of neural manifolds}
To measure the relationship between encoding performance and representational complexity, we compute the \emph{intrinsic dimensionality} $I_d$ as well as the linear \emph{effective dimensionality} $d$ of activations at each layer. $I_d$ and $d$ describe different geometric properties of the representations: while the former describes the dimension of the representations' underlying (nonlinear) manifold, the latter describes the number of linear directions that explain their variance up to a threshold. When unambiguous, we will use the word \emph{dimensionality} to refer to both $I_d$ and $d$, specifying when necessary.

We are interested in an LM's behavior on a representative sample of natural language, so that the computed dimensionality is informative about the LM's linguistic processing in general. For all LMs mentioned in the previous section, we compute the ID on $N=10000$ 20-word contexts randomly sampled from The Pile \citep{gao2020pile}, which constitute Pythia's training data,\footnote{The training data for OPT are not publicly downloadable.} for 5 random data samples. Each sequence is first transformed into a sequence of tokens, which are the atomic (subword) units that constitute the LM's vocabulary. Then, as the tokenization scheme may result in sequences of variable length, we aggregate representations at each layer by taking that of the last token in the model's \emph{residual stream} \citep{elhage2021mathematical}; this yields one $N \times D$ matrix of representations per layer, $D$ being the model's hidden dimension, or extrinsic dimension.

\paragraph{Nonlinear ID estimation} To compute $I_d$, we apply the Generalized Ratios Intrinsic Dimension Estimator (GRIDE) \citep{Denti_Doimo_Laio_Mira_2022}, an extension of the popular TwoNN  estimator \citep{Facco_d’Errico_Rodriguez_Laio_2017} to general scales. GRIDE operates on ratios $\mu_{i, 2k, k} := r_{i, 2k} / r_{i,k}$, where $r_{i,j}$ is the Euclidean distance between point $i$ and its $j^{th}$ neighbor. Assuming local uniform density up to the $2k^{th}$ neighbor, the ratios $\mu_{i, 2k, k}$ follow a generalized Pareto distribution $f_{\mu_i, 2k, k}(\mu) = \frac{I_d(\mu^{I_d} - 1)^{k-1}}{B(k,k)\mu^{I_d(2k-1)+1}}$, where $B(\cdot, \cdot)$ is the beta function. The $I_d$ is then recovered by maximizing this likelihood over points $i$ for several candidate scales $k$. 

Finally, in order to choose the proper $I_d$, a scale analysis over $k$, which controls the neighborhood size, is necessary: if $k$ is too small, the $I_d$ likely describes local noise, and if $k$ is too large, the curvature of the manifold will produce a faulty estimate. Instead, it is recommended to choose a $k$ for which the $I_d$ is stable \citep{Denti_Doimo_Laio_Mira_2022}. We provide an example of such a scale analysis in \Cref{app:id_estimation}.

\paragraph{Linear dimensionality estimation} In addition to nonlinear $I_d$, we computed linear effective dimensionality $d$ two ways: using PCA with variance cutoff $0.99$ \citep{pca}, and the Participation Ratio (PR), defined as $(\sum_i \lambda_i)^2 / (\sum_i \lambda_i ^2)$ \citep{Gao2017ATO}. By definition, $I_d$-dimensional manifolds can be embedded in $d \geq I_d$ dimensions, so we expect that $d \geq I_d$.

\subsection{Measuring layerwise surprisal}
Causal LLMs are trained to predict the next token in context. The LLM produces a conditional probability distribution $p_{LM}(\cdot | x_{<i})$ over the next token $x_i$ given a linguistic context $x_{<i}$. The loss is given by the negative log-likelihood of the ground-truth token under $p_{LM}$, equivalent to minimizing the information-theoretic coding length, or surprisal, of the next token given context. 

This predictive coding objective is hypothesized to underlie sentence processing in humans, and has been proposed to explain brain-LLM representational similarity. To determine whether this is the case, we computed the next-token surprisal from intermediate layers using TunedLens \citep{Belrose2023ElicitingLP} on The Pile dataset. TunedLens learns an affine mapping from an intermediate layer to the vocabulary space in order to predict the next token, indicating how much intermediate layers (linearly) represent next-token identity. See \Cref{app:tunedlens} for implementation details.

\begin{table}[]
    \centering
    \begin{tabular}{r|cccc}
            & OPT-125m & OPT-1.3b & OPT-13b & Pythia-6.9b \\
        GRIDE $I_d$ &  0.91 & \textbf{0.96} & 0.85 & \textbf{0.90} \\
        PCA $d$ &  0.91 & 0.93 & \textbf{0.96} & 0.86 \\
        PR $d$ & \textbf{0.94} & 0.82  &  0.85 & $-0.05^*$
    \end{tabular}
    \caption{The average voxelwise product-moment correlations between representational dimensionality and encoding performance are shown for $I_d$, PCA-$d$ (variance threshold of $0.99$), and PR-$d$. Across models, the correlation is generally high no matter the dimensionality measure. All values, except those marked with (*), are significant to $p < 10^{-3}$, as computed by a permutation test.}
    \label{tab:correlation_dim_ep}
\end{table}

\section{Results}

Layerwise encoding performance and layerwise representational dimensionality across layers are highly correlated, consistently across brain areas involved in linguistic processing. \Cref{tab:correlation_dim_ep} shows the correlation between encoding performance and dimensionality, averaged over all voxels. The relationship is largely consistent across a variety of metrics for measuring dimensionality.

\Cref{fig:phase_transition_lm}a shows the correlation between average encoding performance and normalized $I_d$ for various model sizes in from the OPT model family. The positive relationship, $\rho = 0.85$, between normalized $I_d$ and encoding performance suggests that in trained language models, the $I_d$ of layer activations captures abstract linguistic feature complexity needed to support language comprehension.

\Cref{fig:phase_transition_lm}b overlays, for OPT-1.3b, the encoding performance, $I_d$, and next-token prediction loss computed from each layer. Observe that encoding performance peaks at layer 17, which exactly marks the sharp downwards turn in prediction loss. While \citet{cheng2024emergencehighdimensionalabstractionphase} show that layers leading up to the $I_d$ peak extract high-level features related to syntax and semantics, our results additionally indicate a shift in post-$I_d$-peak to next-token prediction. This sharp phase transition from abstraction to prediction is observed across model sizes, but it is more gradual for Pythia (see \Cref{app:two_phase_abstraction_other_models}). 

To further verify the existence of a phase transition, we report the inter-layer representational similarity via linear Centered Kernel Alignment \citep{kornblith2019similarity}. \Cref{fig:phase_transition_lm}d depicts at least two phases of inter-layer similarity (lighter is more similar): the $I_d$ peak approximately marks a junction at which preceding layers are no longer similar to following layers. Results generally hold across models, see \Cref{app:cka_other_models}.

\Cref{fig:phase_transition_lm}c shows the correlation of $I_d$ with encoding performance across layers at the voxelwise level (red is better), in a single subject. With the exception of the primary auditory cortex, which processes low-level auditory information, encoding performance in brain areas thought to handle higher-level linguistic processing is well-predicted by $I_d$ across layers. Results generally hold across subjects, model families, and model sizes, see \Cref{app:id_voxel_correlation_results}.

\begin{figure}
    \centering
    \includegraphics[width=1\linewidth]{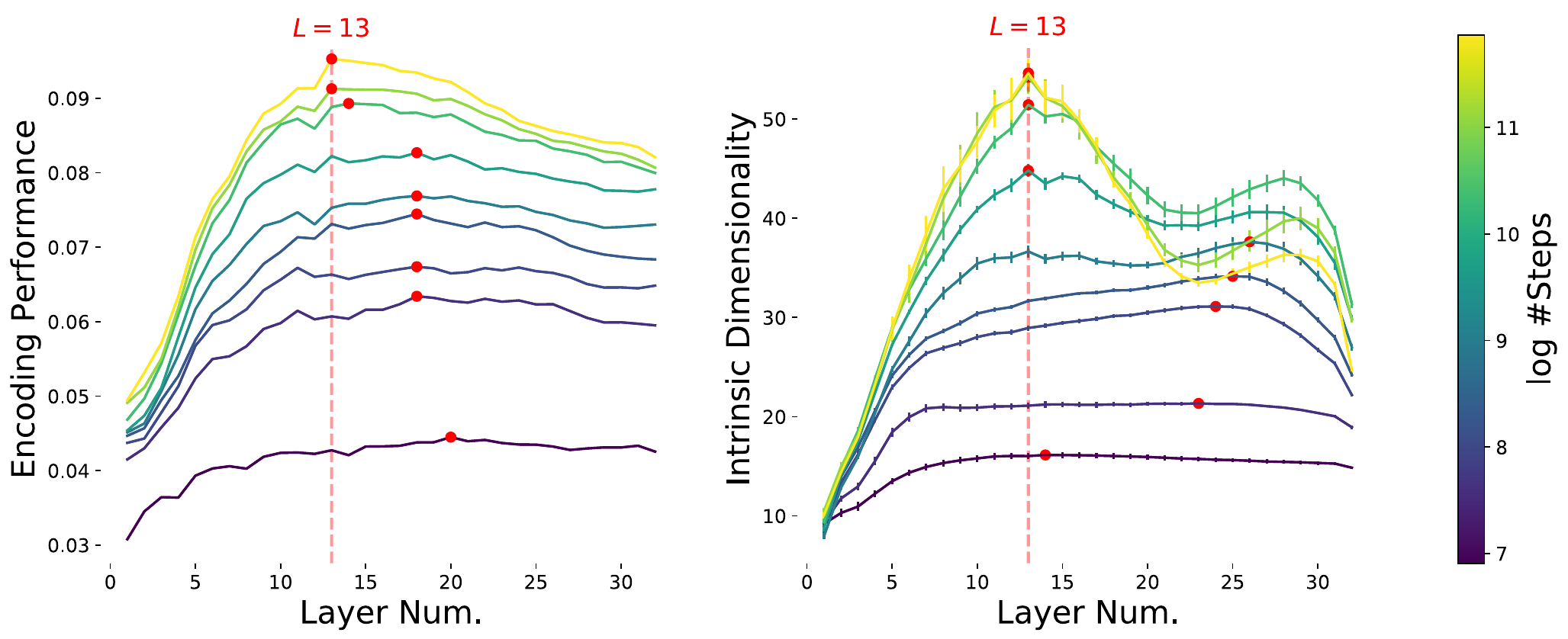}
    \caption{\textit{Encoding Performance and Intrinsic Dimensionality Peaks Manifest Concurrently over Training}: \textbf{(a)} - The evolution of layerwise encoding performance over training of the Pythia 6.9B model is shown. A peak is reached at layer 13 of the model. \textbf{(b)}  -Likewise, a peak in $I_d$ at layer 13 manifests over training. Red dots in each figure denote maximal layers for the respective metric.}
    \label{fig:encoding_id_over_training}
\end{figure}

The relationship between encoding performance and $I_d$ arises nontrivially from learning, in a way that does not simply reflect the layer position. \Cref{fig:encoding_id_over_training} plots the encoding performance and $I_d$ across layers over the course of training for Pythia-6.9B (each curve is a different checkpoint). We confirm an existing result from the literature that the characteristic $I_d$ peak emerges, and moreover, that $I_d$ generally grows for all layers over training (\Cref{fig:encoding_id_over_training} right) \citep{cheng2024emergencehighdimensionalabstractionphase}. Furthermore, the encoding performance (left) and $I_d$ (right) increase at similar rates over training, seen by similar positions of the checkpoint curves in the two plots. The two plots are globally correlated with $\rho = 0.94$. 

The location of the $I_d$ peak (red dots, right), changes over the course of training, eventually settling at the same layers for peak encoding performance (red dots, left). This rules out the possibility that the $I_d$ peak is a trivial function of the Transformer architecture, e.g., layer index.

\section{Discussion}

Recent studies of the properties of language encoding models have observed that the intermediate layers of LLMs, rather than the output layers, have the highest linear similarity to measured brain activity. This is true regardless of the scanning modality (be it fMRI~\cite{NEURIPS2023_4533e4a3}, ECoG~\cite{hong2024scale}, or MEG~\cite{caucheteux2023evidence}), and regardless of the chosen LLM. Despite this very frequently observed trend, little research has been dedicated to explaining this phenomenon. Yet, an understanding of this trend would greatly benefit our understanding of both brains and LLMs, not least because layerwise differences in LLMs have highly useful epistemic properties. LLM layers are invariant to many confounding variables - each layer has seen the same data in the same order, has an identical architecture, was trained on the same loss term, and built using the same hyperparameters. Therefore, differences between layers can only arise either as a result of the compositional nature of the transition from earlier layers to later ones, or due to the "time pressure" exerted by the loss term on the final output layers.

These competing pressures, to first build up the most comprehensive representation of the input text possible, and to then ultimately use this representation to resolve towards a distribution over predicted next word outputs, have opposite effects, as we demonstrate here. The composition effect leads to a increase in encoding performance and dimensionality, whereas the prediction effect narrows the dimensionality to the detriment of encoding. Furthermore, we observe that as models get larger and more thoroughly trained, the best layer for encoding slowly drifts to earlier in the model, perhaps suggesting a saturation effect for this initial compositional phase. 

What conclusions should we draw from this? Firstly, that it is not likely to be the autoregressive nature of language models that drives brain-model similarity~\cite{schrimpf2021neural, goldstein2022shared, antonello2022predictive}. As models get more potent at prediction, their most predictive and most descriptive layers drift apart.\footnote{We note that it is inherently challenging to draw affirmative conclusions about underlying mechanisms in the brain from from brain-model similarity alone. See \citet{guest_martin_2021,antonello2022predictive} for further details.} Secondly, we can draw that the multi-phase abstraction process in LLMs that has been proposed independently by other authors~\cite{cheng2024emergencehighdimensionalabstractionphase,valeriani2023} is supported by evidence from the only other system known to effectively reason with complex language, the human brain. As the present work only tests two model families, it will be necessary to test more models for conclusions to hold in the general case. 

From a practical perspective, conclusions point to a potential new avenue for improving the performance of encoding models. If the spectral properties of different LLM layers can be measured and efficiently combined to produce a representation with higher $I_d$ than any individual layer, then we might expect that new representation to outperform any single layerwise encoding model coming from the same LLM. As linear layerwise encoding models reach their limit, such methods may be necessary to see further benefits. 





\section*{Acknowledgements}
This project has received funding from the European Research Council (ERC) under the European Union’s Horizon 2020 research and innovation programme (grant agreement No. 101019291) as well as the Dingwall Foundation and a computing gift from the Texas Advanced Computing Center (TACC) at The University of Texas at Austin. This paper reflects the authors’ view only, and the funding agency is not responsible for any use that may be made of the information it contains.

The authors would like to thank Alexander Huth, Marco Baroni, Alessandro Laio, and members of the COLT group at Universitat Pompeu Fabra for feedback. The authors additionally thank the organizers of the Brains, Minds, and Machines summer course, where the project was started.

\bibliography{emily}

\begin{thebibliography}{27}
\providecommand{\natexlab}[1]{#1}
\providecommand{\url}[1]{\texttt{#1}}
\expandafter\ifx\csname urlstyle\endcsname\relax
  \providecommand{\doi}[1]{doi: #1}\else
  \providecommand{\doi}{doi: \begingroup \urlstyle{rm}\Url}\fi

\bibitem[Goldstein et~al.(2022)Goldstein, Zada, Buchnik, Schain, Price, Aubrey, Nastase, Feder, Emanuel, Cohen, et~al.]{goldstein2022shared}
Ariel Goldstein, Zaid Zada, Eliav Buchnik, Mariano Schain, Amy Price, Bobbi Aubrey, Samuel~A Nastase, Amir Feder, Dotan Emanuel, Alon Cohen, et~al.
\newblock Shared computational principles for language processing in humans and deep language models.
\newblock \emph{Nature neuroscience}, 25\penalty0 (3):\penalty0 369--380, 2022.

\bibitem[Vaidya et~al.(2022)Vaidya, Jain, and Huth]{vaidya2022self}
Aditya~R Vaidya, Shailee Jain, and Alexander~G Huth.
\newblock Self-supervised models of audio effectively explain human cortical responses to speech.
\newblock \emph{arXiv preprint arXiv:2205.14252}, 2022.

\bibitem[Jain et~al.(2023)Jain, Vo, Wehbe, and Huth]{jain2023computational}
Shailee Jain, Vy~A Vo, Leila Wehbe, and Alexander~G Huth.
\newblock Computational language modeling and the promise of in silico experimentation.
\newblock \emph{Neurobiology of Language}, pages 1--65, 2023.

\bibitem[Antonello et~al.(2023)Antonello, Vaidya, and Huth]{NEURIPS2023_4533e4a3}
Richard Antonello, Aditya Vaidya, and Alexander Huth.
\newblock Scaling laws for language encoding models in fmri.
\newblock In A.~Oh, T.~Naumann, A.~Globerson, K.~Saenko, M.~Hardt, and S.~Levine, editors, \emph{Advances in Neural Information Processing Systems}, volume~36, pages 21895--21907. Curran Associates, Inc., 2023.
\newblock URL \url{https://proceedings.neurips.cc/paper_files/paper/2023/file/4533e4a352440a32558c1c227602c323-Paper-Conference.pdf}.

\bibitem[Tuckute et~al.(2023)Tuckute, Sathe, Srikant, Taliaferro, Wang, Schrimpf, Kay, and Fedorenko]{tuckute2023driving}
Greta Tuckute, Aalok Sathe, Shashank Srikant, Maya Taliaferro, Mingye Wang, Martin Schrimpf, Kendrick Kay, and Evelina Fedorenko.
\newblock Driving and suppressing the human language network using large language models.
\newblock \emph{bioRxiv}, 2023.

\bibitem[OOTA et~al.(2023)OOTA, Gupta, and Toneva]{NEURIPS2023_3a0e2de2}
SUBBAREDDY OOTA, Manish Gupta, and Mariya Toneva.
\newblock Joint processing of linguistic properties in brains and language models.
\newblock In A.~Oh, T.~Naumann, A.~Globerson, K.~Saenko, M.~Hardt, and S.~Levine, editors, \emph{Advances in Neural Information Processing Systems}, volume~36, pages 18001--18014. Curran Associates, Inc., 2023.
\newblock URL \url{https://proceedings.neurips.cc/paper_files/paper/2023/file/3a0e2de215bd17c39ad08ba1d16c1b12-Paper-Conference.pdf}.

\bibitem[Mischler et~al.(2024)Mischler, Li, Bickel, Mehta, and Mesgarani]{mischler2024contextual}
Gavin Mischler, Yinghao~Aaron Li, Stephan Bickel, Ashesh~D Mehta, and Nima Mesgarani.
\newblock Contextual feature extraction hierarchies converge in large language models and the brain.
\newblock \emph{arXiv preprint arXiv:2401.17671}, 2024.

\bibitem[Caucheteux et~al.(2023)Caucheteux, Gramfort, and King]{caucheteux2023evidence}
Charlotte Caucheteux, Alexandre Gramfort, and Jean-R{\'e}mi King.
\newblock Evidence of a predictive coding hierarchy in the human brain listening to speech.
\newblock \emph{Nature Human Behaviour}, pages 1--12, 2023.

\bibitem[Schrimpf et~al.(2021)Schrimpf, Blank, Tuckute, Kauf, Hosseini, Kanwisher, Tenenbaum, and Fedorenko]{schrimpf2021neural}
Martin Schrimpf, Idan~Asher Blank, Greta Tuckute, Carina Kauf, Eghbal~A Hosseini, Nancy Kanwisher, Joshua~B Tenenbaum, and Evelina Fedorenko.
\newblock The neural architecture of language: Integrative modeling converges on predictive processing.
\newblock \emph{Proceedings of the National Academy of Sciences}, 118\penalty0 (45):\penalty0 e2105646118, 2021.

\bibitem[Antonello and Huth(2022)]{antonello2022predictive}
Richard Antonello and Alexander Huth.
\newblock Predictive coding or just feature discovery? an alternative account of why language models fit brain data.
\newblock \emph{Neurobiology of Language}, pages 1--16, 2022.

\bibitem[Valeriani et~al.(2023)Valeriani, Doimo, Cuturello, Laio, Ansuini, and Cazzaniga]{valeriani2023}
Lucrezia Valeriani, Diego Doimo, Francesca Cuturello, Alessandro Laio, Alessio Ansuini, and Alberto Cazzaniga.
\newblock The geometry of hidden representations of large transformer models.
\newblock In A.~Oh, T.~Naumann, A.~Globerson, K.~Saenko, M.~Hardt, and S.~Levine, editors, \emph{Advances in Neural Information Processing Systems}, volume~36, pages 51234--51252. Curran Associates, Inc., 2023.
\newblock URL \url{https://proceedings.neurips.cc/paper_files/paper/2023/file/a0e66093d7168b40246af1cddc025daa-Paper-Conference.pdf}.

\bibitem[Cheng et~al.(2024)Cheng, Doimo, Kervadec, Macocco, Yu, Laio, and Baroni]{cheng2024emergencehighdimensionalabstractionphase}
Emily Cheng, Diego Doimo, Corentin Kervadec, Iuri Macocco, Jade Yu, Alessandro Laio, and Marco Baroni.
\newblock Emergence of a high-dimensional abstraction phase in language transformers, 2024.
\newblock URL \url{https://arxiv.org/abs/2405.15471}.

\bibitem[Schrimpf et~al.(2018)Schrimpf, Kubilius, Hong, Majaj, Rajalingham, Issa, Kar, Bashivan, Prescott-Roy, Geiger, et~al.]{schrimpf2018brain}
Martin Schrimpf, Jonas Kubilius, Ha~Hong, Najib~J Majaj, Rishi Rajalingham, Elias~B Issa, Kohitij Kar, Pouya Bashivan, Jonathan Prescott-Roy, Franziska Geiger, et~al.
\newblock Brain-score: Which artificial neural network for object recognition is most brain-like?
\newblock \emph{BioRxiv}, page 407007, 2018.

\bibitem[Belrose et~al.(2023)Belrose, Furman, Smith, Halawi, Ostrovsky, McKinney, Biderman, and Steinhardt]{Belrose2023ElicitingLP}
Nora Belrose, Zach Furman, Logan Smith, Danny Halawi, Igor~V. Ostrovsky, Lev McKinney, Stella Biderman, and Jacob Steinhardt.
\newblock Eliciting latent predictions from transformers with the tuned lens.
\newblock \emph{ArXiv}, abs/2303.08112, 2023.
\newblock URL \url{https://api.semanticscholar.org/CorpusID:257504984}.

\bibitem[Zhang et~al.(2022)Zhang, Roller, Goyal, Artetxe, Chen, Chen, Dewan, Diab, Li, Lin, Mihaylov, Ott, Shleifer, Shuster, Simig, Koura, Sridhar, Wang, and Zettlemoyer]{Zhang_Roller_Goyal_Artetxe_Chen_Chen_Dewan_Diab_Li_Lin_et}
Susan Zhang, Stephen Roller, Naman Goyal, Mikel Artetxe, Moya Chen, Shuohui Chen, Christopher Dewan, Mona Diab, Xian Li, Xi~Victoria Lin, Todor Mihaylov, Myle Ott, Sam Shleifer, Kurt Shuster, Daniel Simig, Punit~Singh Koura, Anjali Sridhar, Tianlu Wang, and Luke Zettlemoyer.
\newblock Opt: Open pre-trained transformer language models.
\newblock \penalty0 (arXiv:2205.01068), Jun 2022.
\newblock \doi{10.48550/arXiv.2205.01068}.
\newblock URL \url{http://arxiv.org/abs/2205.01068}.
\newblock arXiv:2205.01068 [cs].

\bibitem[Biderman et~al.(2023)Biderman, Schoelkopf, Anthony, Bradley, O’Brien, Hallahan, Khan, Purohit, Prashanth, Raff, et~al.]{biderman2023pythia}
Stella Biderman, Hailey Schoelkopf, Quentin~Gregory Anthony, Herbie Bradley, Kyle O’Brien, Eric Hallahan, Mohammad~Aflah Khan, Shivanshu Purohit, USVSN~Sai Prashanth, Edward Raff, et~al.
\newblock Pythia: A suite for analyzing large language models across training and scaling.
\newblock In \emph{International Conference on Machine Learning}, pages 2397--2430. PMLR, 2023.

\bibitem[Gao et~al.(2020)Gao, Biderman, Black, Golding, Hoppe, Foster, Phang, He, Thite, Nabeshima, et~al.]{gao2020pile}
Leo Gao, Stella Biderman, Sid Black, Laurence Golding, Travis Hoppe, Charles Foster, Jason Phang, Horace He, Anish Thite, Noa Nabeshima, et~al.
\newblock The {P}ile: An 800{GB} dataset of diverse text for language modeling.
\newblock \emph{arXiv preprint arXiv:2101.00027}, 2020.

\bibitem[Elhage et~al.(2021)Elhage, Nanda, Olsson, Henighan, Joseph, Mann, Askell, Bai, Chen, Conerly, DasSarma, Drain, Ganguli, Hatfield-Dodds, Hernandez, Jones, Kernion, Lovitt, Ndousse, Amodei, Brown, Clark, Kaplan, McCandlish, and Olah]{elhage2021mathematical}
Nelson Elhage, Neel Nanda, Catherine Olsson, Tom Henighan, Nicholas Joseph, Ben Mann, Amanda Askell, Yuntao Bai, Anna Chen, Tom Conerly, Nova DasSarma, Dawn Drain, Deep Ganguli, Zac Hatfield-Dodds, Danny Hernandez, Andy Jones, Jackson Kernion, Liane Lovitt, Kamal Ndousse, Dario Amodei, Tom Brown, Jack Clark, Jared Kaplan, Sam McCandlish, and Chris Olah.
\newblock A mathematical framework for transformer circuits.
\newblock \emph{Transformer Circuits Thread}, 2021.
\newblock https://transformer-circuits.pub/2021/framework/index.html.

\bibitem[Denti et~al.(2022)Denti, Doimo, Laio, and Mira]{Denti_Doimo_Laio_Mira_2022}
Francesco Denti, Diego Doimo, Alessandro Laio, and Antonietta Mira.
\newblock The generalized ratios intrinsic dimension estimator.
\newblock \emph{Scientific Reports}, 12\penalty0 (11):\penalty0 20005, Nov 2022.
\newblock ISSN 2045-2322.
\newblock \doi{10.1038/s41598-022-20991-1}.

\bibitem[Facco et~al.(2017)Facco, d’Errico, Rodriguez, and Laio]{Facco_d’Errico_Rodriguez_Laio_2017}
Elena Facco, Maria d’Errico, Alex Rodriguez, and Alessandro Laio.
\newblock Estimating the intrinsic dimension of datasets by a minimal neighborhood information.
\newblock \emph{Scientific Reports}, 7\penalty0 (1):\penalty0 12140, Sep 2017.
\newblock ISSN 2045-2322.
\newblock \doi{10.1038/s41598-017-11873-y}.

\bibitem[Jolliffe(1986)]{pca}
Ian Jolliffe.
\newblock \emph{Principal Component Analysis}.
\newblock Springer, 1986.

\bibitem[Gao et~al.(2017)Gao, Trautmann, Yu, Santhanam, Ryu, Shenoy, and Ganguli]{Gao2017ATO}
Peiran Gao, Eric~M. Trautmann, Byron~M. Yu, Gopal Santhanam, Stephen~I. Ryu, Krishna~V. Shenoy, and Surya Ganguli.
\newblock A theory of multineuronal dimensionality, dynamics and measurement.
\newblock \emph{bioRxiv}, 2017.
\newblock URL \url{https://api.semanticscholar.org/CorpusID:19938440}.

\bibitem[Kornblith et~al.(2019)Kornblith, Norouzi, Lee, and Hinton]{kornblith2019similarity}
Simon Kornblith, Mohammad Norouzi, Honglak Lee, and Geoffrey Hinton.
\newblock Similarity of neural network representations revisited.
\newblock In \emph{International Conference on Machine Learning}, pages 3519--3529. PMLR, 2019.

\bibitem[Hong et~al.(2024)Hong, Wang, Zada, Gazula, Turner, Aubrey, Niekerken, Doyle, Devore, Dugan, et~al.]{hong2024scale}
Zhuoqiao Hong, Haocheng Wang, Zaid Zada, Harshvardhan Gazula, David Turner, Bobbi Aubrey, Leonard Niekerken, Werner Doyle, Sasha Devore, Patricia Dugan, et~al.
\newblock Scale matters: Large language models with billions (rather than millions) of parameters better match neural representations of natural language.
\newblock \emph{bioRxiv}, pages 2024--06, 2024.

\bibitem[Guest and Martin(2021)]{guest_martin_2021}
Olivia Guest and Andrea~E Martin.
\newblock On logical inference over brains, behaviour, and artificial neural networks, 10 2021.
\newblock URL \url{psyarxiv.com/tbmcg}.

\bibitem[Ghandeharioun et~al.(2024)Ghandeharioun, Caciularu, Pearce, Dixon, and Geva]{ghandeharioun2024patchscope}
Asma Ghandeharioun, Avi Caciularu, Adam Pearce, Lucas Dixon, and Mor Geva.
\newblock Patchscope: A unifying framework for inspecting hidden representations of language models.
\newblock \emph{arXiv preprint arXiv:2401.06102}, 2024.

\bibitem[LeBel et~al.(2022)LeBel, Wagner, Jain, Adhikari-Desai, Gupta, Morgenthal, Tang, Xu, and Huth]{lebel2022natural}
Amanda LeBel, Lauren Wagner, Shailee Jain, Aneesh Adhikari-Desai, Bhavin Gupta, Allyson Morgenthal, Jerry Tang, Lixiang Xu, and Alexander~G Huth.
\newblock A natural language fmri dataset for voxelwise encoding models.
\newblock \emph{bioRxiv}, pages 2022--09, 2022.

\end{thebibliography}
\bibliographystyle{unsrtnat}


\appendix
\renewcommand\thefigure{\thesection.\arabic{figure}}    
\renewcommand\thetable{\thesection.\arabic{table}}
\renewcommand\theequation{\thesection.\arabic{equation}}

\section{Computing resources}
\label{app:computing-resources}
\setcounter{figure}{0}    
\setcounter{table}{0} 

Dimensionality and surprisal computation were run on a cluster with 12 nodes with 5 NVIDIA A30 GPUs and 48 CPUs each. Extracting and computing dimensionality on LM representations took a few wall-clock hours per model. Training TunedLens took around 15 minutes per layer, so overall 30 wall-clock hours. We parallelized all computation, and estimate the overall parallelized runtime, including preliminary experiments and failed runs to be around 6 days.

Ridge regression was performed using compute nodes with 128 cores (2 AMD EPYC 7763 64-core processors) and 256GB of RAM. In total, roughly 1,000 node-hours of compute was expended for these models. Feature extraction for language models was performed on specialized GPU nodes similar to the AMD compute nodes but with 3 NVIDIA A100 40GB cards. Feature extraction required roughly 300 node-hours of compute on these GPU nodes. 

\section{ID Estimation}
\label{app:id_estimation}
\setcounter{figure}{0}    
\setcounter{table}{0} 

\begin{figure}
    \centering
    \includegraphics[width=0.75\linewidth]{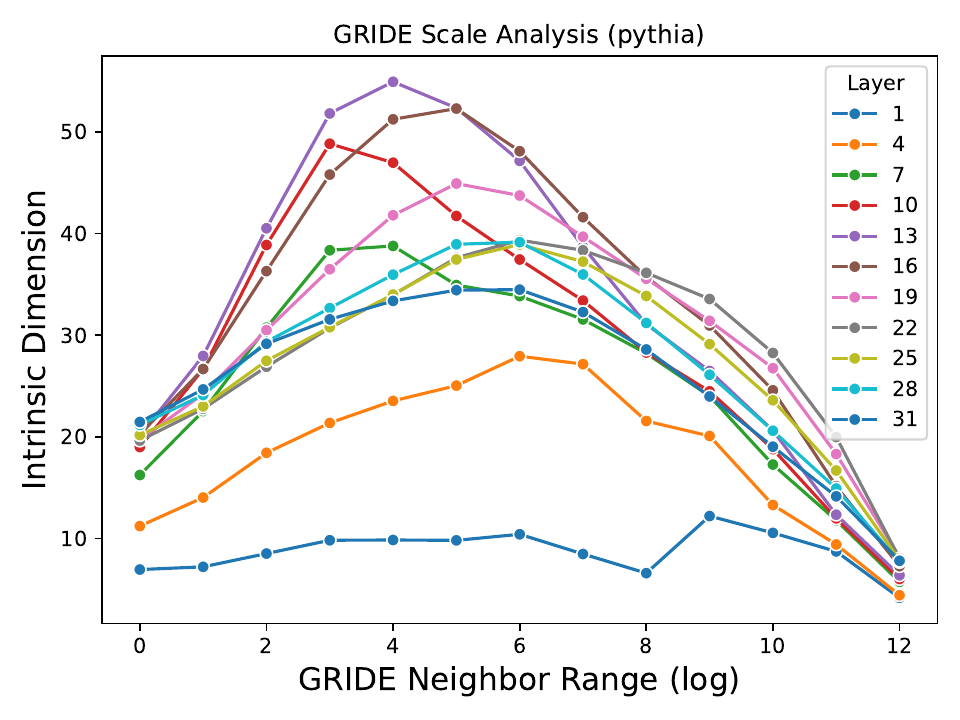}
    \caption{GRIDE scale analysis for Pythia-6.9b. The estimated intrinsic dimension (y axis) varies according to the chosen scale $k$ (x axis). It is recommended to choose a scale where the local change is minimal, in this case, $k=2^4$.}
    \label{fig:scale_analysis}
\end{figure}

For ID estimation using GRIDE, we reproduce the setup in \citet{cheng2024emergencehighdimensionalabstractionphase}. For each model, checkpoint, and layer, we perform a scale analysis. The intrinsic dimension of the manifold is sensitive to the \emph{scale}, or neighborhood size, for which it is estimated \citep{Facco_d’Errico_Rodriguez_Laio_2017,Denti_Doimo_Laio_Mira_2022}. \Cref{fig:scale_analysis} shows an example, where the GRIDE scale $k$ varies from $2^0$ to $2^{12}$. As recommended in \citet{Denti_Doimo_Laio_Mira_2022}, we choose a scale $k$ corresponding in a range where the intrinsic dimension is stable, or plateaus, by visual inspection. For simplicity, we choose one scale $k$ per model. In the particular example in \Cref{fig:scale_analysis}, we choose $k=2^4$, where the derivative of the curve is closest to 0 for as many layers as possible. Scales chosen for all models are in \Cref{tab:scales}. 

\begin{table}[]
    \centering
    \begin{tabular}{l|l}
        Model & GRIDE $k$ \\
        \hline 
        OPT-125m & 64 \\
        OPT-1.3b & 32 \\
        OPT-13b & 32 \\
        Pythia-6.9B & 16 \\
        Pythia ($t=$64000) & 16 \\
        Pythia ($t=$32000) & 32 \\
        Pythia ($t=$16000) & 32 \\
        Pythia ($t=$8000) & 32 \\
        Pythia ($t=$4000) & 64 \\
        Pythia ($t=$3000) & 64 \\
        Pythia ($t=$2000) & 16 \\
        Pythia ($t=$1000) & 16 \\
        Pythia ($t=$512) & 16 \\ 
    \end{tabular}
    \caption{Selected GRIDE scales $k$ after performing a scale analysis for intrinsic dimension estimation, for all models and checkpoints tested.}
    \label{tab:scales}
\end{table}

\section{Surprisal Estimation}
\label{app:tunedlens}
\setcounter{figure}{0}    
\setcounter{table}{0} 

We used the TunedLens implementation by \citet{ghandeharioun2024patchscope}. TunedLens ascertains the amount of information (linearly) encoded in hidden layer $t$ about the next token. To do so, an affine mapping is learned from the last-token hidden representation $h_t$ at layer $t$ as follows:

\begin{equation}
    \min_{A_t, b_t} \mathcal D_{KL}(f_{>t}(h_t)\ \ || \ \ \text{LayerNorm}(A_t h_t + b_t)W_U).
\end{equation}

Here, $A_t \in \mathbb R^{D\times D}$, $b_t \in \mathbb R^D$ are the learnable parameters of the affine mapping. $W_U$ is the LM's unembedding matrix that maps the final layer to the vocabulary. Finally, $f_{>t}(h_t)$ is the layers of the LM $f$ after layer $t$, producing the model's original distribution over the vocabulary. In the provided code \citep{ghandeharioun2024patchscope}, TunedLens is implemented using a direct solver $\texttt{numpy.linalg.lstsq}$ on $N=8000$ randomly sampled sequences from The Pile dataset \citep{gao2020pile}, returning the least squares solution that minimizes the $l_2$-norm between $h_t$ and last layer representation $h_T$. Finally, we compute the next-token surprisals on a validation set of The Pile ($N=2000$) from the TunedLens-modified hidden layers. 

\section{fMRI Methods}
\label{app:fmri}

 MRI data were collected on a 3T Siemens Skyra scanner at The University of Texas at Austin Biomedical Imaging Center using a 64-channel Siemens volume coil. Functional scans were collected using a gradient echo EPI sequence with repetition time (TR) = 2.00 s, echo time (TE) = 30.8 ms, flip angle = 71°, multi-band factor (simultaneous multi-slice) = 2, voxel size = 2.6mm x 2.6mm x 2.6mm (slice thickness = 2.6mm), matrix size = 84x84, and field of view = 220 mm. Anatomical data were collected using a T1-weighted multi-echo MP-RAGE sequence with voxel size = 1mm x 1mm x 1mm.

In addition to motion correction and coregistration \cite{lebel2022natural}, low frequency voxel response drift was identified using a 2nd order Savitzky-Golay filter with a 120 second window and then subtracted from the signal. The mean response for each voxel was subtracted and the remaining response was scaled to have unit variance.

\clearpage

\section{Extended Results}
\setcounter{figure}{0}    
\setcounter{table}{0} 

\subsection{Extended Tuned Lens Results}
\label{app:two_phase_abstraction_other_models}
\begin{figure}[h]
    \centering
    \includegraphics[width=1\textwidth]{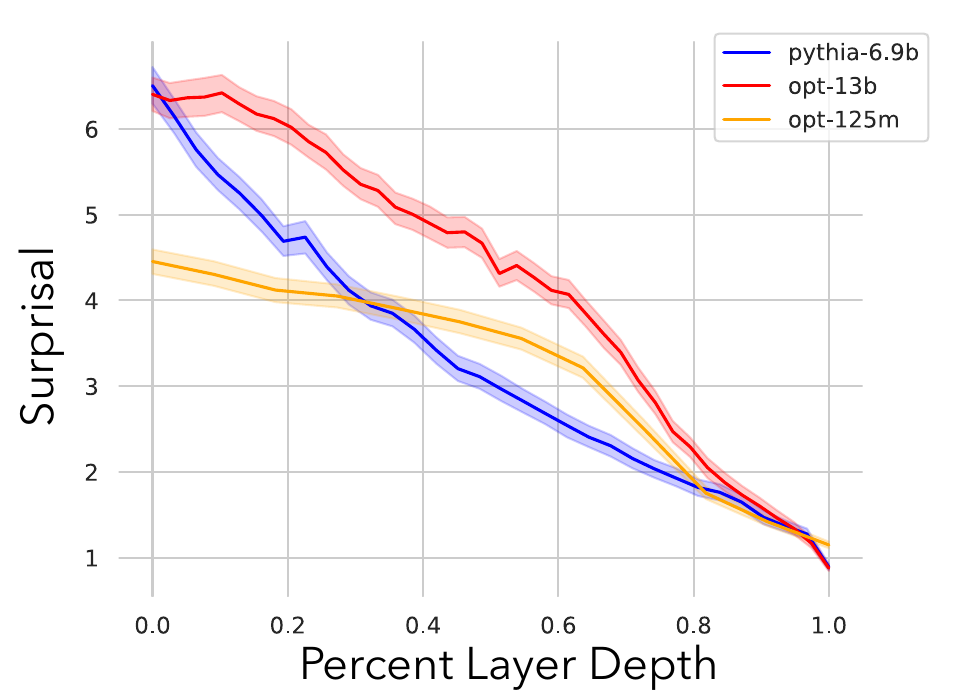}
    \caption{Remaining tuned lens results for OPT-125, OPT-13B, and Pythia-6.9B}
    \label{fig:pythia-cka}
\end{figure}

\clearpage

\subsection{Extended Voxelwise ID Correlation Results}
\label{app:id_voxel_correlation_results}

\begin{figure}[h]
    \centering
    \includegraphics[width=1\textwidth]{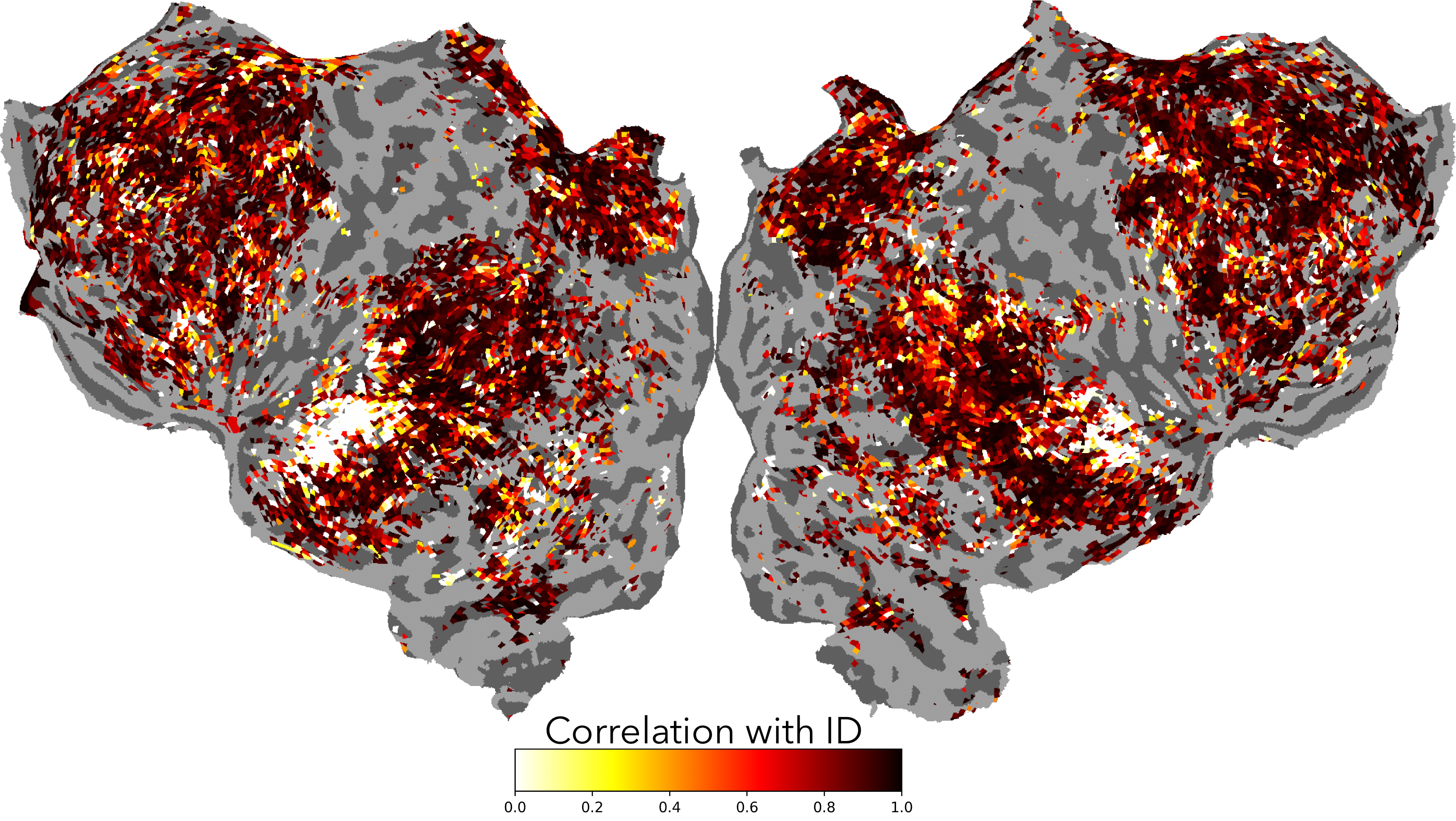}
    \caption{Voxelwise ID correlation results as in Figure 1c for OPT-125M}
    \label{fig:pythia-cka}
\end{figure}

\begin{figure}[h]
    \centering
    \includegraphics[width=1\textwidth]{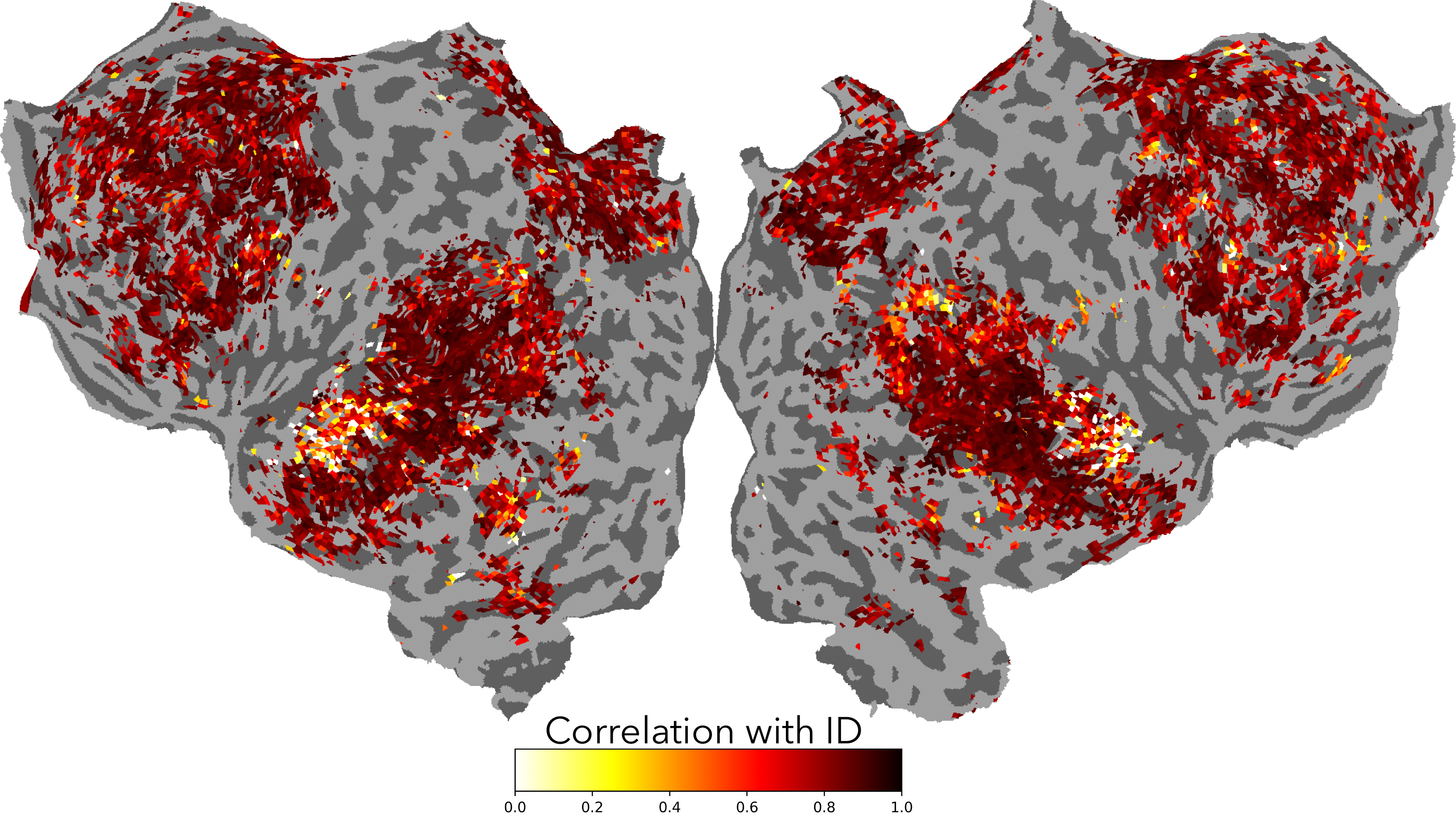}
    \caption{Voxelwise ID correlation results as in Figure 1c for OPT-13B}
    \label{fig:pythia-cka}
\end{figure}

\begin{figure}[h]
    \centering
    \includegraphics[width=1\textwidth]{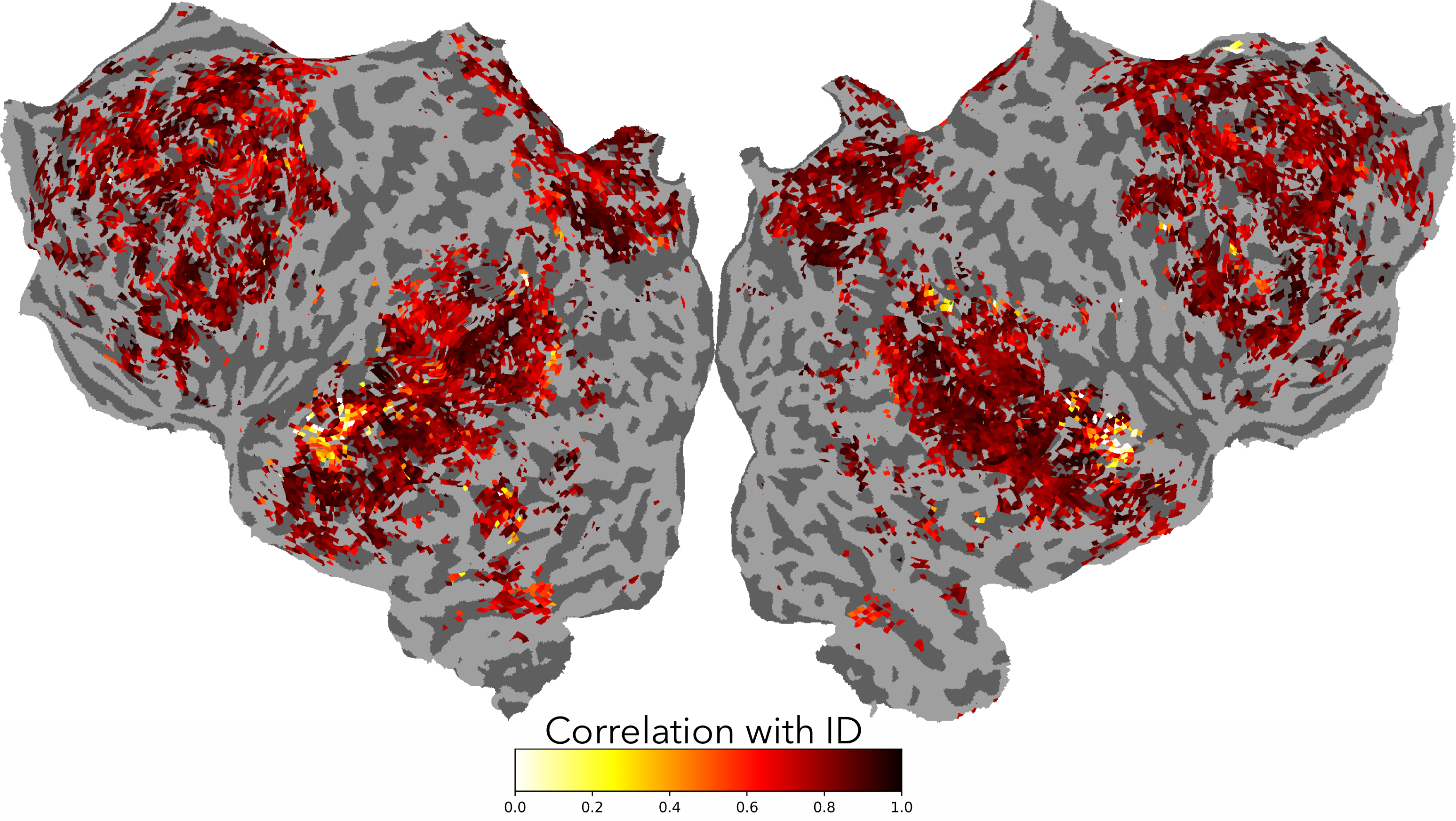}
    \caption{Voxelwise ID correlation results as in Figure 1c for Pythia-6.9B}
    \label{fig:pythia-cka}
\end{figure}

\clearpage

\subsection{Extended CKA Results}
\label{app:cka_other_models}

\begin{figure}[h]
    \centering
    \includegraphics[width=0.7
\textwidth]{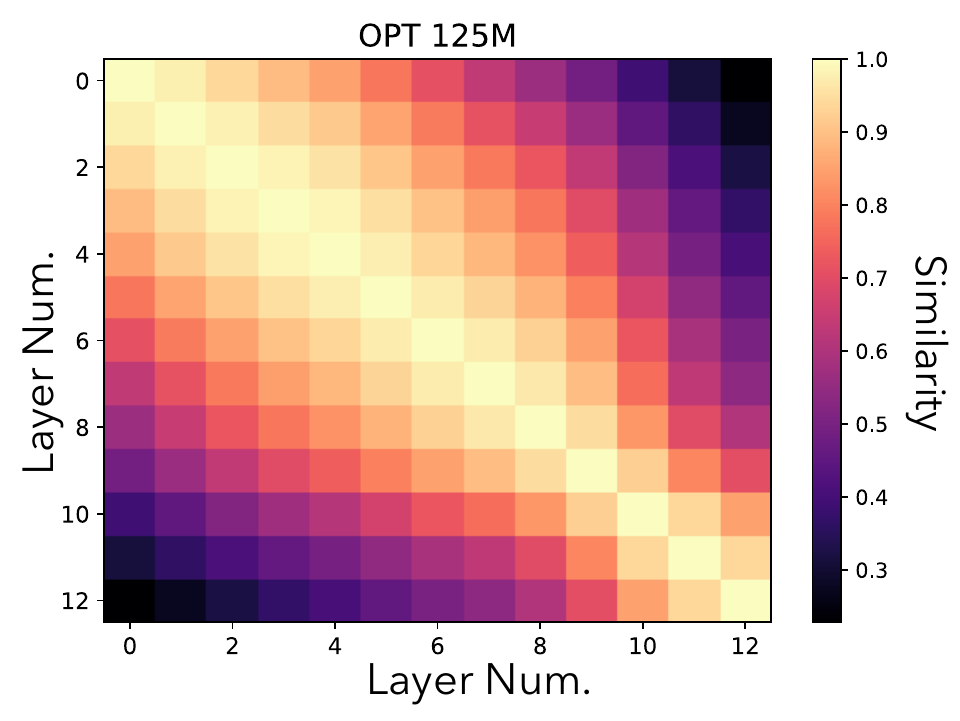}
    \caption{CKA results as in Figure 1d for OPT-125M}
    \label{fig:opt125m-cka}
\end{figure}

\begin{figure}[h]
    \centering
    \includegraphics[width=1\textwidth]{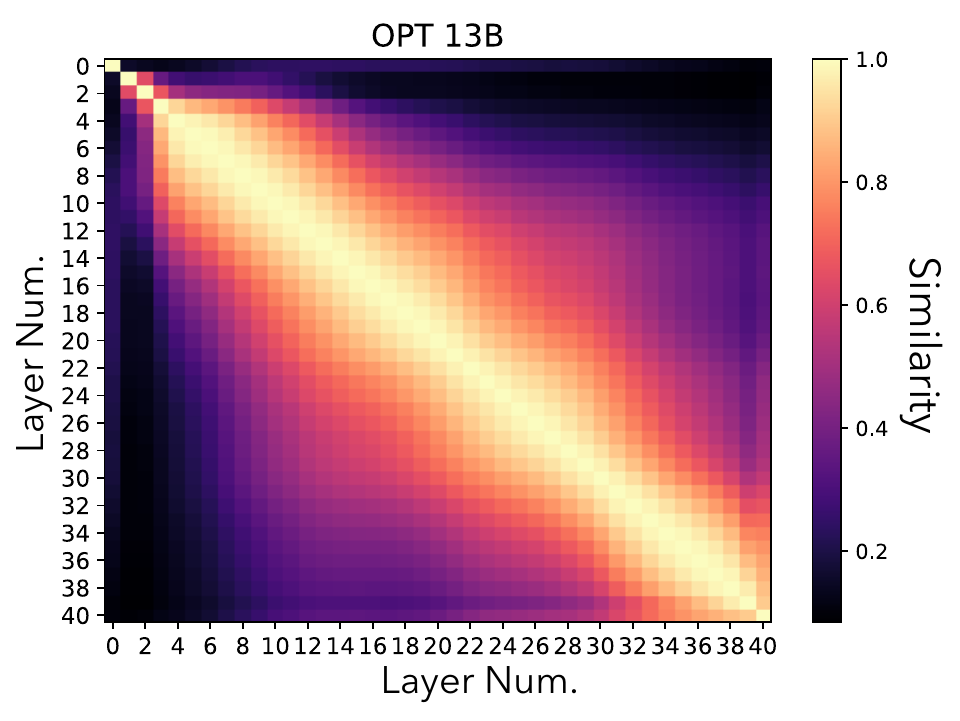}
    \caption{CKA results as in Figure 1d for OPT-13B}
    \label{fig:opt13b-cka}
\end{figure}

\begin{figure}[h]
    \centering
    \includegraphics[width=1\textwidth]{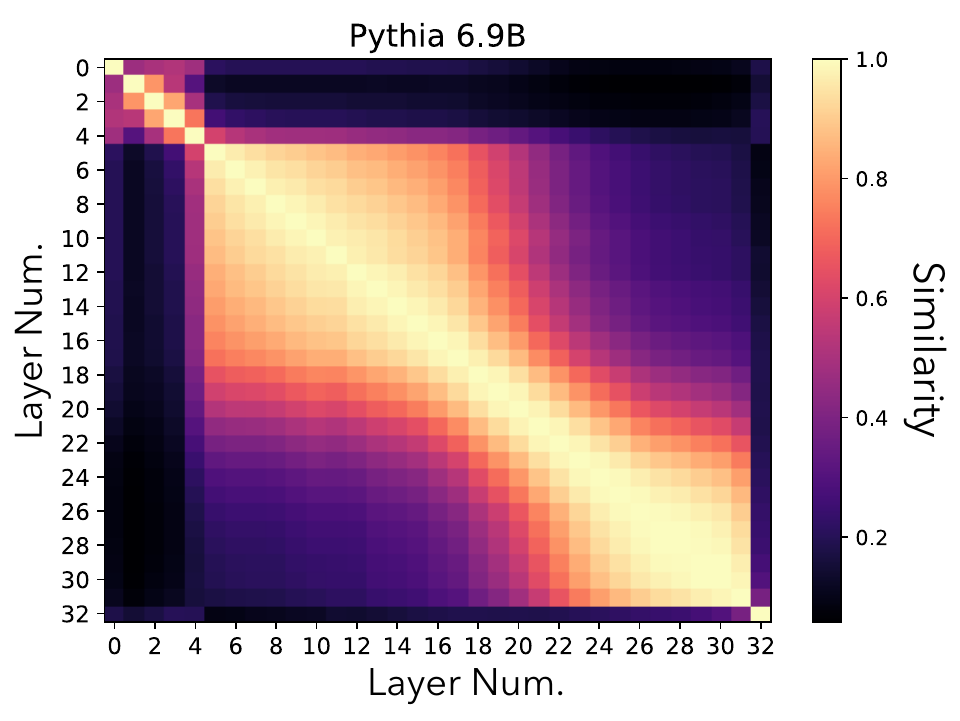}
    \caption{CKA results as in Figure 1d for Pythia-6.9B}
    \label{fig:pythia-cka}
\end{figure}

\end{document}